\pdfoutput=1 

\documentclass[10pt]{article} 

\usepackage[accepted]{rlj} 

\usepackage{amssymb}            
\usepackage{mathtools}          
\usepackage{mathrsfs}           
\usepackage{graphicx}           
\usepackage{subcaption}         
\usepackage[space]{grffile}     
\usepackage{url}                

\usepackage[utf8]{inputenc}     
\usepackage{booktabs}           
\usepackage{amsfonts}           
\usepackage{nicefrac}           
\usepackage{microtype}          
\usepackage{bbm}
\usepackage{dsfont}
\usepackage{multirow}
\usepackage{algorithm}
\usepackage{algorithmic}

\usepackage{amsmath,amsfonts,bm}

\def\eqref#1{equation~\ref{#1}}

\def\1{\bm{1}}

\DeclareMathAlphabet{\mathsfit}{\encodingdefault}{\sfdefault}{m}{sl}
\SetMathAlphabet{\mathsfit}{bold}{\encodingdefault}{\sfdefault}{bx}{n}

\newcommand{\bd}[1]{\textcolor{green}{[BD: #1]}}

\title{PB\textsuperscript{2}: Preference Space Exploration via Population-Based Methods in Preference-Based Reinforcement Learning}

\setrunningtitle{PB\textsuperscript{2}: Population-Based Preference-Based RL}

\author{Brahim Driss\textsuperscript{1}, Alex Davey\textsuperscript{1,2}, Riad Akrour\textsuperscript{1}}

\emails{\{brahim.driss,alex.davey,riad.akrour\}@inria.fr}

\affiliations{
$^{1}$\textbf{Univ. Lille, Inria, CNRS, Centrale Lille, UMR 9189 -- CRIStAL, Lille, France}\\
$^{2}$\textbf{Inria TAU, LISN, Université Paris-Saclay, Orsay, France}
}

\contribution{
    We identify the preference exploration problem in PbRL, showing how single-agent methods can converge to suboptimal local minima in the preference space.
    }
    {
    Existing PbRL query selection methods \citep{Wang2023RUNE, Nakamoto2023QPA, Biyik2020} all operate within single-agent frameworks. This is an empirical observation, not a formal impossibility result.
    }

\contribution{
    We propose PB\textsuperscript{2}, a population-based framework for PbRL that maintains policy diversity while optimizing for human preferences, broadening coverage of the preference space.
    }
    {
    PB\textsuperscript{2} adapts performance-constrained diversity from SMERL \citep{Kumar2020SMERL} to PbRL, where no expert policy is available, and can be viewed as an instantiation of dueling posterior sampling \citep{dps2020} with an explicit diversity bonus.
    }

\contribution{
    We show experimentally that neural ensemble reward modeling \citep{Christiano2017, Lee2021b, Wang2023RUNE} induces little behavioral diversity, justifying the need for an explicit diversity mechanism.
    }
    {
    Most PbRL methods rely on reward ensembles to capture preference uncertainty, but we observe these ensembles lose diversity quickly. This compares a Thompson Sampling baseline against single-agent QPA \citep{Nakamoto2023QPA}; the observation is empirical.
    }

\contribution{
    We demonstrate experimentally that PB\textsuperscript{2} produces more distinguishable queries, improving reward learning efficiency with greater robustness to inconsistent human feedback and limited feedback budgets.
    }
    {
    All experiments use simulated feedback following the B-Pref protocol \citep{Lee2021a}, with a similarity threshold introducing inconsistent labels when trajectory returns are close. Results are reported on DMControl and navigation tasks.
    }


\summary{Preference-based reinforcement learning (PbRL) enables agents to learn behaviors from human feedback without requiring predefined reward functions. However, existing PbRL methods struggle to explore the preference space effectively, often converging prematurely to suboptimal policies that capture only a narrow subset of human preferences. In this work, we identify and address this preference exploration problem through population-based methods. By maintaining a diverse population of agents, our approach achieves broader coverage of the preference space than single-agent methods. This diversity yields two benefits: it improves reward model learning, and it generates preference queries with clearly distinguishable behaviors, reducing ambiguity for human evaluators. Our experiments show that existing methods can become trapped in local optima, requiring excessive feedback. Their performance also degrades substantially when evaluators make errors on similar trajectories, a realistic condition that methods assuming perfect oracle feedback tend to overlook. In contrast, our population-based approach explores preferences more effectively in environments with complex reward structures and is considerably more robust to noisy evaluators.}

\begin{document}
\raggedbottom

\makeCover  
\maketitle  

\begin{abstract}
 Preference-based reinforcement learning (PbRL) enables agents to learn behaviors from human feedback without requiring predefined reward functions. However, existing PbRL methods struggle to explore the preference space effectively, often converging prematurely to suboptimal policies that capture only a narrow subset of human preferences. In this work, we
identify and address this preference exploration problem through population-based methods. By maintaining a diverse population of agents, our approach achieves broader coverage of the preference space than single-agent methods. This diversity yields two benefits: it improves reward model learning, and it generates preference queries with clearly distinguishable behaviors, reducing ambiguity for human evaluators. Our experiments show that existing methods can become trapped in local optima, requiring excessive feedback. Their performance also degrades substantially when evaluators make errors on similar trajectories, a realistic condition that methods assuming perfect oracle feedback tend to overlook. In contrast, our population-based approach explores preferences more effectively in environments with complex reward structures and is considerably more robust to noisy evaluators.
\end{abstract}

\section{Introduction}
Reinforcement learning (RL) has been applied successfully to game playing, robotic control, and other domains. However, traditional RL methods depend on carefully designed reward functions, which are often difficult to specify for tasks involving subjective outcomes or complex human preferences \citep{HadfieldMenell2017}. Preference-based reinforcement learning (PbRL) addresses this by letting agents learn directly from human feedback through pairwise comparisons of behavior trajectories \citep{Christiano2017, Lee2021a}, removing the need for hand-crafted reward functions. Despite this advantage, PbRL faces a persistent problem: policies optimized for the current reward model tend to generate similar queries for preference elicitation, limiting the informativeness of human feedback needed to improve the model.


Existing approaches attempt to address preference elicitation challenges through various query selection strategies: uncertainty-based methods targeting uncertain preference regions \citep{VARIQuery}, policy-aligned techniques focusing on current behavior \citep{Nakamoto2023QPA}, ensemble-based strategies using model disagreement \citep{Wang2023RUNE}, and information-theoretic approaches maximizing information gain \citep{Biyik2020, biyik2018batch}. While improving sample efficiency, these methods operate within a single-agent framework, limiting behavioral diversity during evaluation. This becomes particularly problematic when humans must compare similar trajectories, as evaluators often provide inconsistent feedback in such cases \citep{huang2025trend}, adding noise that degrades learning performance. Moreover, one of the main claims of this paper is that existing reward modeling approaches based on neural ensembles~\citep{Christiano2017,Lee2021b,Wang2023RUNE} in practice induce little behavioral diversity and might be a poor representation of the rewards' posterior distribution. Without an explicit mechanism for maintaining diversity, existing approaches remain vulnerable to local optima---requiring excessive feedback to recover---and suffer severe degradation in performance under realistic conditions of human evaluation inconsistency.

\begin{figure}[t]
    \centering
    \includegraphics[width=0.75\linewidth]{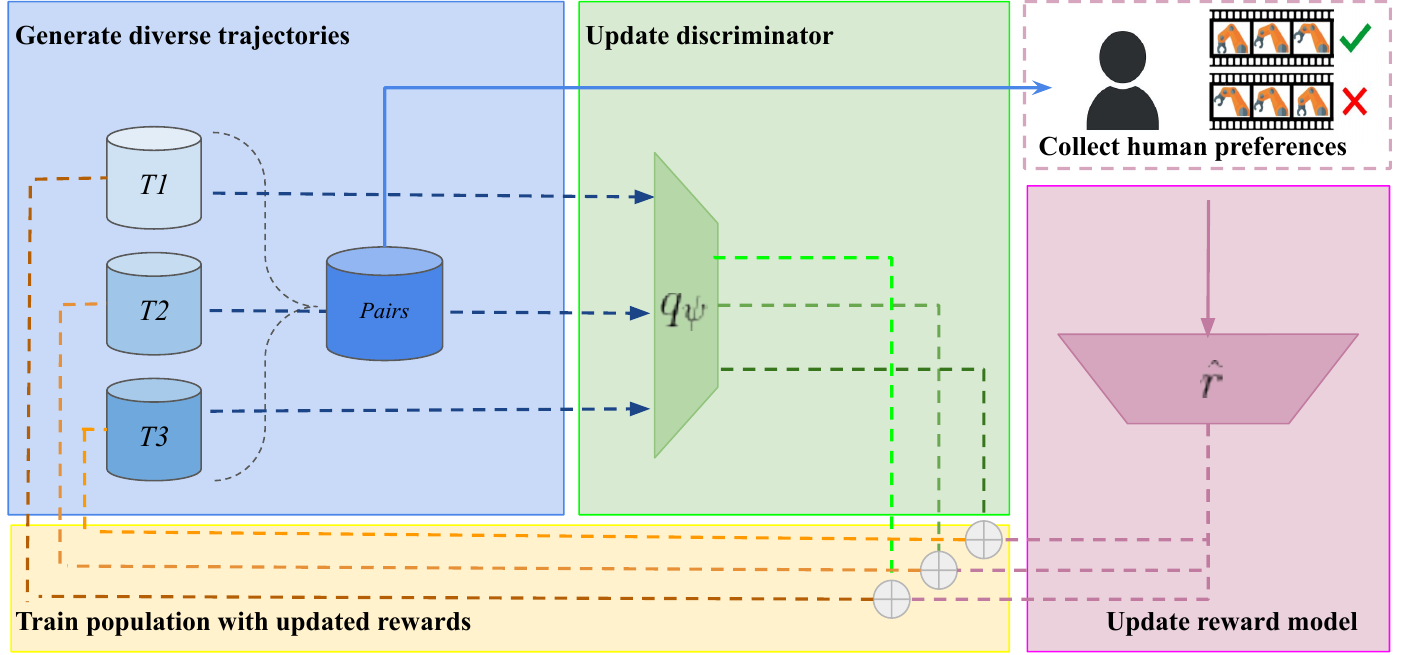}
    \caption{\textbf{Overview of PB²}. Agents in the population optimize a combination of a reward learned from user queries and a bonus term from a discriminator trained to maintain agent distinguishability. This combined reward encourages the discovery of diverse behaviors, while maintaining alignment with human preferences, thereby helping to propose informative and less ambiguous user queries.}
    \label{fig:overview}
\end{figure}

To improve on this lack of behavioral diversity during the user elicitation step of PbRL, we present in this paper PB²: a population-based approach to preference space exploration. The central idea is that simultaneously training multiple distinct policies with an explicit diversity bonus allows broader preference space exploration than conventional single-policy methods. By collecting experiences across different policies to construct comparison pairs, we broaden the variety of behaviors evaluated while preserving alignment with expressed human preferences. As illustrated in Fig.\ref{fig:overview}, diverse policies generate distinct trajectories, human preferences on these trajectories train a reward model, and a discriminator maintains population diversity while encouraging behaviors that align with current preferences. Our contributions are as follows:
\begin{enumerate}
   \item We identify the preference exploration problem in PbRL, demonstrating how single-agent methods can converge to suboptimal local minima in the preference space.
   \item We propose PB\textsuperscript{2}, a population-based framework for PbRL that maintains policy diversity while optimizing for human preferences, broadening coverage of the preference space.
   \item We show experimentally that neural ensemble reward modeling induces little behavioral diversity, performing similarly to not modeling reward uncertainty at all, justifying the need for an explicit diversity mechanism.
   \item We demonstrate experimentally that PB\textsuperscript{2} produces more distinguishable queries, improving reward learning efficiency with greater robustness to inconsistent human feedback and limited feedback budgets.
\end{enumerate}

We validate these claims through three complementary experimental evaluations: (1) a systematic evaluation across DMControl locomotion tasks with varying similarity thresholds $\epsilon$ to simulate human judgment inconsistency; (2) a qualitative demonstration of how PB² escapes local optima in complex preference spaces where single-agent methods remain trapped; and (3) a comparative analysis in navigation tasks with extremely limited feedback, showing PB²'s feedback efficiency.


\section{Related Work}
\textbf{Preference-based Reinforcement Learning.} Preference-based Reinforcement Learning (PbRL) enables agents to learn from human feedback through trajectory comparisons \citep{Christiano2017, Lee2021a, Wirth2017}. A persistent exploration difficulty in PbRL is that policies optimized for the current reward model tend to produce similar queries, reducing the informativeness of subsequent human feedback. Recent methods address this through reward uncertainty \citep{Wang2023RUNE}, semi-supervised learning~\citep{Park2022SURF}, model-based approaches \citep{Li2022MOPRL}, bi-level optimization \citep{Chen2022MRN}, dynamics encoding \citep{Shen2023REED}, and query-policy alignment~(QPA, \citep{Nakamoto2023QPA}), though they remain primarily exploitative. RIME \citep{cheng2024rime} addresses a complementary challenge of noisy feedback through learning to detect and handle mislabeling errors.

Recent work has also explored different strategies for improving query selection and exploration. PPE \citep{PPE2024} improves preference buffer coverage through proximal policy exploration guided by out-of-distribution detection and a mixture distribution query method that balances exploration of new trajectories with maintaining reward model accuracy on in-distribution data. SENIOR \citep{SENIOR2025} combines motion-distinction-based query selection with preference-guided exploration using intrinsic rewards, though operating in offline settings. While these methods improve various aspects of PbRL, they operate within single-agent frameworks that limit behavioral diversity during query generation.

Bayesian approaches like Dueling Posterior Sampling~(DPS, \cite{dps2020}) generate queries from policies maximizing posterior samples of the reward function, providing strong theoretical guarantees for tackling the exploration/exploitation of user preferences. Our approach can be thought of as an instantiation of DPS, but we show in this paper that a naive implementation thereof without an explicit diversity bonus is in fact not much different than the purely exploitative QPA algorithm that only optimizes the maximum likelihood reward with no attempt at reward uncertainty modeling.

\textbf{Population-based Reinforcement Learning and Quality-Diversity.} Population-based methods maintain multiple agents to enhance exploration, from Population-Based Training \citep{Jaderberg2017} to extensions incorporating Bayesian optimization \citep{Wan2022BGPBT}, long-term performance \citep{Dalibard2021FIREPBT}, and evolutionary selection \citep{Salimans2017, Alam2020}. Quality Diversity algorithms discover diverse high-performing solutions, evolving from Novelty Search \citep{Lehman2011a} through approaches like NSLC \citep{Lehman2011b} and MAP-Elites \citep{Mouret2015}, with recent integration into RL \citep{Nilsson2021PGAMAPELITES, Cully2019AURORA}. Our approach shares core principles with these methods but targets preference space exploration rather than hyperparameter optimization, building on diversity-promoting concepts like DvD \citep{NEURIPS2020_d1dc3a82} while aligning diversity with human preferences. Unlike QD methods that require manually designed descriptors, our method automatically identifies meaningful behavioral patterns and adapts to evolving human preferences rather than pursuing general diversity.

\textbf{Unsupervised Skill Discovery.} Skill discovery leverages intrinsic motivation through mutual information maximization \citep{Eysenbach2019}, dynamics-awareness \citep{Sharma2020}, and task-aligned methods \citep{Kumar2020SMERL, osa2022discovering}. Our approach is inspired by SMERL \citep{Kumar2020SMERL} but operates without expert access, encouraging distinctiveness between behaviors from different population members rather than between latent-conditioned policies of a single agent. However, despite these advances in diversity promotion across different domains, these methods have not yet been applied in PbRL to maintain behavioral diversity during query generation, which we show to be a challenge for current methods in Sec. \ref{sec:motiv}.

\section{Preliminaries}
\label{sec4}
We review the standard RL framework and then introduce preference-based learning.

\textbf{Reinforcement Learning.}
We consider the standard reinforcement learning framework with an environment modeled as a Markov Decision Process (MDP) $M := (S, A, T, r, \gamma)$ \citep{sutton1998reinforcement}, where $S$ is the state space, $A$ is the action space, $T$ is the transition function, $r$ is the reward function, and $\gamma$ is the discount factor. A trajectory $\tau = (s_0, a_0, s_1, a_1, \ldots)$ represents a sequence of state-action pairs generated by following policy $\pi$, and the return $R(\tau) = \sum_{t=0}^{\infty} \gamma^t r(s_t, a_t)$ quantifies the total discounted reward obtained along this trajectory. The agent's goal is to find a policy $\pi$ that maximizes the expected discounted cumulative reward, $\mathbb{E}_\pi[\sum_{t=0}^{\infty} \gamma^t r(s_t, a_t)]$.

\textbf{Preference-Based Reinforcement Learning.}
In preference-based reinforcement learning (PbRL) \citep{Christiano2017, Wirth2017}, the agent lacks access to an explicit reward function. Instead, the learning process relies on human feedback provided through pairwise comparisons. During training, the algorithm generates queries by selecting pairs of trajectory segments $(\sigma^0, \sigma^1)$ from the agent's experience, where a segment $\sigma$ is a sequence of state-action pairs extracted from collected trajectories. These segment pairs are presented to a human teacher, who provides preference feedback by indicating which segment better aligns with the desired behavior. The preference label $y$ is a binary indicator: $y^{(1)} = 1$ if the human prefers $\sigma^1$ over $\sigma^0$, and $y^{(0)} = 1$ otherwise.

Following the Bradley-Terry model \citep{bradley1952rank, furnkranz2012preference}, we learn a reward function $r_\phi$ that induces a preference predictor:
\begin{equation}
P_\phi[\sigma^1 \succ \sigma^0] = \frac{\exp\left(\sum_{t=1}^{T} \gamma^{t-1} r_\phi(s^1_t, a^1_t)\right)}{\sum_{i\in\{0,1\}} \exp\left(\sum_{t=1}^{T} \gamma^{t-1} r_\phi(s^i_t, a^i_t)\right)}
\end{equation}

The reward function parameters $\phi$ are optimized by minimizing the cross-entropy loss over collected preference data $D$:
\begin{equation}
L_{\text{reward}} = -\mathbb{E}_{(\sigma^0, \sigma^1, y) \sim D}\left[y^{(0)}\log P_\phi[\sigma^0 \succ \sigma^1] + y^{(1)}\log P_\phi[\sigma^1 \succ \sigma^0]\right]
\end{equation}

This learned reward function then guides policy training through standard RL algorithms. The reward function parameters $\phi$ are shared across all agents in the population, providing a common objective while agents maintain distinct behaviors through the diversity mechanism described in Section 5.1. However, the effectiveness of this approach depends strongly on how queries are generated and selected.

\begin{figure}[t]
    \centering
    \includegraphics[width=\linewidth]{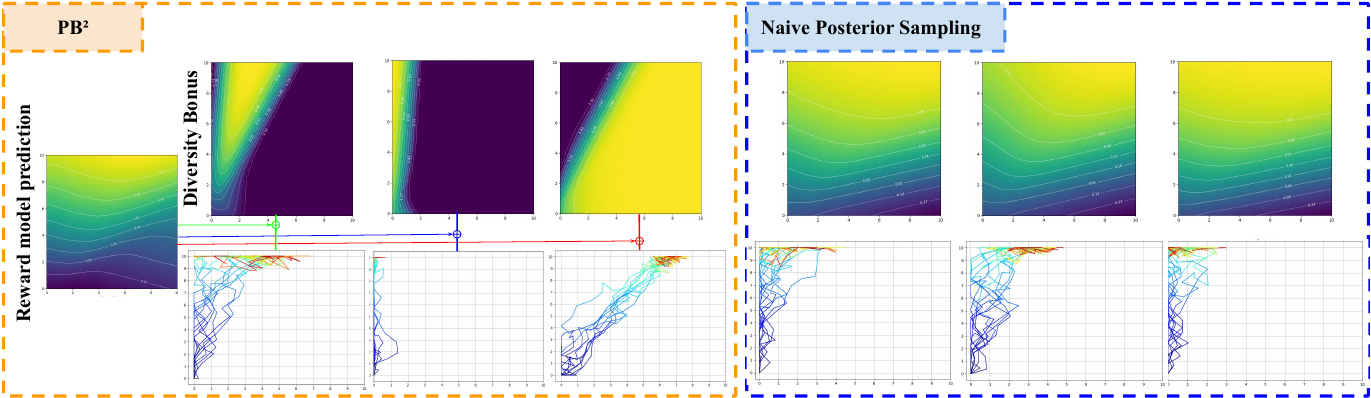}
    \caption{\textbf{Diverse population-based query strategy improves preference space coverage.}  2D navigation task with a goal on the top right corner. \textit{\textbf{Orange box}} shows the current reward model that prefers states in the upper region, and  diversity bonuses (Sec. \ref{sec:diversity}) for three agents, with distinct spatial concentrations. Bottom row shows agent trajectories maximizing the combination of the learned reward model and the diversity bonus, except for the first agent that always only maximizes the current reward model in PB². This agent corresponds to what QPA would learn. On itself, this agent only generates a moderate amount of behavioral diversity stemming from its policy entropy, while the population of PB² collectively covers a wider region of the state space while staying aligned with human preferences. This desirable property is not as clear without an explicit diversity bonus as shown in the \textit{\textbf{Blue box}}. Here agents maximize their individual reward models (top row) of a learned neural ensemble following standard PbRL procedures. While these reward functions would ideally represent samples of the posterior reward distribution, they end up very similar looking and induce trajectories with a lot more overlap than PB², suggesting that existing reward modeling techniques in PbRL fail to properly represent the uncertainty in user preferences.
    }
    \label{fig:diversity_rewards}
\end{figure}

\section{Balancing Preference Space Exploration and Exploitation}
\label{sec:motiv}
In PbRL, human feedback is expensive and limited. The selection of queries presented to human evaluators directly affects the learning process. PbRL systems must balance two competing objectives: \textbf{exploitation}, which consists in refining the reward model in areas most relevant to current policy improvement, and \textbf{exploration}, which aims to discover previously unknown aspects of human preferences.

Current approaches struggle with this balance. PEBBLE \citep{Lee2021b} samples queries from outdated trajectories, creating temporal misalignment between feedback and current policies. Uncertainty-driven approaches like RUNE \citep{Wang2023RUNE} often prioritize informationally rich regions that may be irrelevant to current capabilities. Even QPA \citep{Nakamoto2023QPA}, which aligns queries with current policies, becomes overly exploitative, creating blind spots in the preference model and potentially trapping agents in local optima.

Counterintuitively, despite extensive research into active query selection, the current PbRL state-of-the-art (QPA) still relies on random query selection from a purely exploitative agent, contradicting fundamental insights from preference elicitation literature \citep{eus2010}. We demonstrate that the underlying problem lies in reward modeling itself: single-agent approaches lack behavioral diversity, leading to increasingly similar trajectory pairs that provide diminishing information. In principle, if reward modeling were sufficiently accurate, explicit diversity terms  might be unnecessary \citep{dps2020}, but this is not the case in practice as shown in Fig. \ref{fig:diversity_rewards}.

Fig.\ref{fig:diversity_rewards} compares preference exploration strategies in a 2D navigation task. Trajectories from a single agent (orange box) concentrate in high-reward regions but provide limited exploration. In contrast, our population generates trajectories that collectively cover a wider area, including informative boundary regions, and are more distinguishable for human evaluators. A population can improve diversity for several reasons: a single neural policy may fail to represent multimodal trajectory distributions, or the primacy bias \citep{nikishin2022primacy} may limit representation capabilities over time. However, the population alone is not enough: without an explicit diversity bonus (blue box), population diversity still collapses because the neural ensemble poorly represents the reward posterior. We test this with a Thompson Sampling baseline (TS): a naive instantiation of DPS that keeps PB²'s population but replaces the explicit diversity bonus with a standard neural ensemble, giving each agent its own reward-model sample \citep{Christiano2017,Lee2021b}. We show in Sec.~\ref{sec:distinguishability} that TS behaves like the purely exploitative QPA, confirming that vanilla neural ensembles poorly capture the reward posterior.

\section{PB²: Population-Based Preference-Based Reinforcement Learning}
\label{sec5}
In this section, we present the components of PB²: a \textbf{P}opulation-\textbf{B}ased approach for \textbf{P}reference-\textbf{B}ased Reinforcement Learning that covers the preference space while maintaining alignment with human preferences. To combat diversity collapse of reward neural ensembles, we employ in PB² an information-theoretic formulation that maximizes the mutual information between policies and their state distributions. This encourages each agent to explore a distinct subset of states that can be readily identified by a discriminator, naturally generating clearly distinguishable behaviors for human evaluation. The challenge is then to maintain diversity in a non-stationary context, since the interactive nature of PbRL creates a continuously evolving reward function, as detailed in the following sections.


\subsection{Performance-Constrained Diversity for Preference Exploration}
\label{sec:diversity}
PB² addresses the preference exploration challenge through a population-based approach with two components: (1) an anchor, a reference policy that tracks achievable performance under current preferences, and (2) diverse policies that explore the preference space while remaining aligned with human intent.

\begin{algorithm}[t]
\caption{PB²: Population-Based Preference-Based RL}
\begin{algorithmic}[1]
\STATE Initialize anchor policy $\pi_{\text{ref}}$, diverse policies $\{\pi_i\}_{i=2}^N$, discriminator $q_\psi$, reward model $r_\phi$ \hfill \emph{(Section 5)}
\STATE Perform initial unsupervised exploration to collect diverse trajectories \hfill \emph{\citep{Lee2021b}}
\WHILE{feedback budget not exhausted}
\STATE Sample trajectories from all policies (anchor + diverse) \hfill \emph{(Section \ref{sec5})}
\STATE Collect human preferences on trajectory segments pairs across policies \hfill \emph{(Fig. \ref{fig:overview}, Section \ref{sec5})}
\STATE Update reward model $r_\phi$ to predict preferences \hfill \emph{(Section \ref{sec4})}
\STATE Update $\pi_{\text{ref}}$ to maximize $r_\phi(\tau)$ only \hfill \emph{(Section \ref{sec:diversity})}
\STATE $R_{\text{ref}} = \mathbb{E}[R_\phi(\tau)]$ for trajectories from $\pi_{\text{ref}}$ \hfill \emph{(Section \ref{sec:diversity})}
\FOR{each diverse policy $\pi_i, i \geq 2$}
  \STATE $R_i = \mathbb{E}[R_\phi(\tau)]$ for trajectories from $\pi_i$ \hfill \emph{(Section \ref{sec:diversity})}

    \IF{$R_i \geq \alpha \cdot R_{\text{ref}}$}
        \STATE Update $\pi_i$ to maximize $r_\phi + \lambda \cdot \log q_\psi(i|.)$ using SAC
        \hfill \emph{(Section \ref{sec:diversity})}
    \ELSE
        \STATE Update $\pi_i$ to maximize $r_\phi$ only using SAC \hfill \emph{(Section \ref{sec:diversity})}
    \ENDIF
\ENDFOR
\STATE Train $q_\psi$ to maximize $\mathbb{E}_{i \sim p(i), s \sim \pi_i}[\log q_\psi(i|s)]$ \hfill \emph{(Section \ref{sec:diversity}, Eq. \ref{eq:disc})}
\ENDWHILE
\end{algorithmic}
\label{alg:pb2}
\end{algorithm}

\textbf{Performance-Constrained Diversity.} Building on the performance-constrained diversity principle introduced in SMERL \citep{Kumar2020SMERL}, we adapt this mechanism to the preference learning setting. Unlike SMERL, which requires access to an expert or optimal return value, PB² operates in settings where the reward function itself is being learned. Similarly to QPA, we learn a single reward model that plays the role of the maximum likelihood reward function. From this model, we generate additional reward `samples' by mixing the maximum likelihood model with diversity bonus functions.

To balance maximization of the learned reward model and the diversity bonus, our approach maintains a reference policy (the anchor policy $\pi_{\text{ref}}$) that purely maximizes the current reward model, establishing a performance baseline, which we denote by $R_{\text{ref}} = \mathbb{E}_{\tau_{\text{ref}} \sim \pi_{\text{ref}}} [R_\phi(\tau_{\text{ref}})]$. The remaining policies are encouraged to develop distinct behaviors through a discriminator $q_\psi$ that predicts which agent generated a given state, but only when their performance is within a specified threshold of the anchor:

\begin{equation}
\pi^*_i \in \arg\max_{\pi_i} \mathbb{E}_{s,a \sim \pi_i} \left[ r_\phi(s,a) + \lambda \cdot \mathds{1}_{[R_i \geq \alpha \cdot R_{\text{ref}}]} \cdot \log q_\psi(i|s) \right]
\end{equation}

where $r_\phi$ is the learned reward function, $R_i = \mathbb{E}[\sum_t \gamma^t r_\phi(s_t,a_t)]$ is agent $i$'s expected return, $\mathds{1}_{[R_i \geq \alpha \cdot R_{\text{ref}}]}$ is an indicator function that equals 1 when the agent's expected return is at least $\alpha$ times the anchor's return, $p(i) = 1/N$ is the uniform prior over the $N$ agents in the population (used in Eq.~\ref{eq:disc}), and $q_\psi$ is a learned discriminator that predicts which agent generated a given state.

\textbf{Adaptive Discriminator.}
To maintain diversity as preferences evolve, we employ a discriminator trained to maximize the mutual information between policies and their state distributions:
\begin{equation}
\mathcal{L}_{\text{disc}}(\psi) = \mathbb{E}_{i \sim p(i), s \sim \pi_i} \left[ \log q_\psi(i|s) \right]
\label{eq:disc}
\end{equation}
\\
Unlike approaches with fixed reward objectives, our discriminator adapts continuously to the evolving behavior distribution. This adaptation is important when human preferences shift, as it encourages policies to discover new distinguishable behaviors that remain aligned with current preferences. When the reward model $r_\phi$ is updated with new preferences, we temporarily disable the diversity bonus, allowing policies to first adapt to the new reward function before reintroducing diversity incentives. This prevents diversity from interfering with the initial adaptation to updated preferences.
Fig.\ref{fig:diversity_rewards} illustrates how this mechanism results in each agent receiving distinct exploration bonuses that collectively span the preference space. This setting differs structurally from standard RL with diversity bonuses (e.g., DIAYN, \citealp{Eysenbach2019}): there the reward is fixed and non-stationarity arises only from the changing state distribution, whereas in PbRL the objective itself changes as $r_\phi$ is re-estimated from new feedback. The anchor agent and the suspension of the diversity bonus after each update are what keep the population aligned as the objective changes, and are unnecessary when the reward is fixed. Implementation details are provided in Appendix~\ref{app:implementation}.

\section{Experiments}
We evaluate PB² across diverse environments to demonstrate its effectiveness in preference space exploration and learning from limited human feedback.

\subsection{Experimental setup}
\textbf{Environments and Baselines.}
We use (1) \textbf{Low-feedback environments} (2D Navigation and PointMaze), where \emph{low-feedback} refers to the small preference-query budget ($N{=}4$ to $20$) rather than reward sparsity, for testing exploration efficiency, and (2) \textbf{DMControl locomotion tasks} \citep{tassa2018deepmind} for evaluating distinguishability. We compare against \textit{PEBBLE} \citep{Lee2021b}, \textit{RUNE} \citep{Wang2023RUNE}, \textit{QPA} \citep{Nakamoto2023QPA}, and \textit{RIME} \citep{cheng2024rime}, representing the primary query selection strategies in PbRL: historical sampling, uncertainty-driven exploration, current-policy alignment, and noise handling. We also compare against a naive dueling posterior sampling baseline TS \citep{dps2020}.

\textbf{Evaluation methodology.} We use the ground truth reward (hidden from algorithms) to evaluate performance, reporting mean and standard deviation across 10 seeds for the DMControl results (Fig.~\ref{fig:dmc_exp}) and 5 seeds for the navigation results (Table~\ref{tab:nav2d_feedback}). For each query, the simulated teacher simply prefers the segment with the higher ground-truth return, so the training preferences are a ranking of the true returns (made noisy by the threshold $\epsilon$ below). The Bradley-Terry model (Eq.~1) enters only when we fit the learned reward $r_\phi$ to these labels; we do not use it to generate the labels themselves. To simulate human inconsistency, we assign random preference labels when trajectory returns are within a similarity threshold $\epsilon$, i.e., $|R_1 - R_2| < \epsilon \cdot \max(R_1, R_2)$, following the Equal teacher from B-Pref \citep{Lee2021a}. We emphasize results with $\epsilon>0$ as more realistic. Full details are in Appendix~\ref{app:eval_methodology}.

\subsection{Exploration efficiency in low-feedback environments}

\begin{figure}[t]
    \centering
    \includegraphics[width=0.68\linewidth]{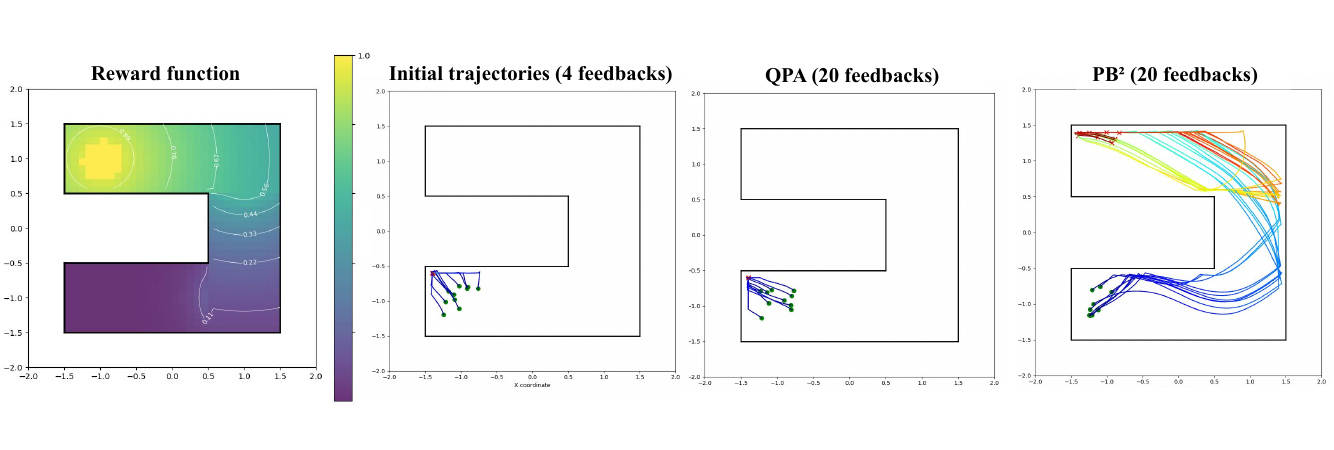}
    \caption{\textbf{Escaping local optima in preference landscapes.} After the same initial feedback (4 queries), both algorithms learn similar reward models favoring the upper-left region. After 20 feedback instances, QPA (middle) stays trapped in this suboptimal region while PB² (right) discovers a path to the high-reward region (left) through its diverse population.}
    \label{fig:qpa_failure}
\end{figure}

\begin{figure}[t]
    \centering
    \includegraphics[width=0.72\linewidth]{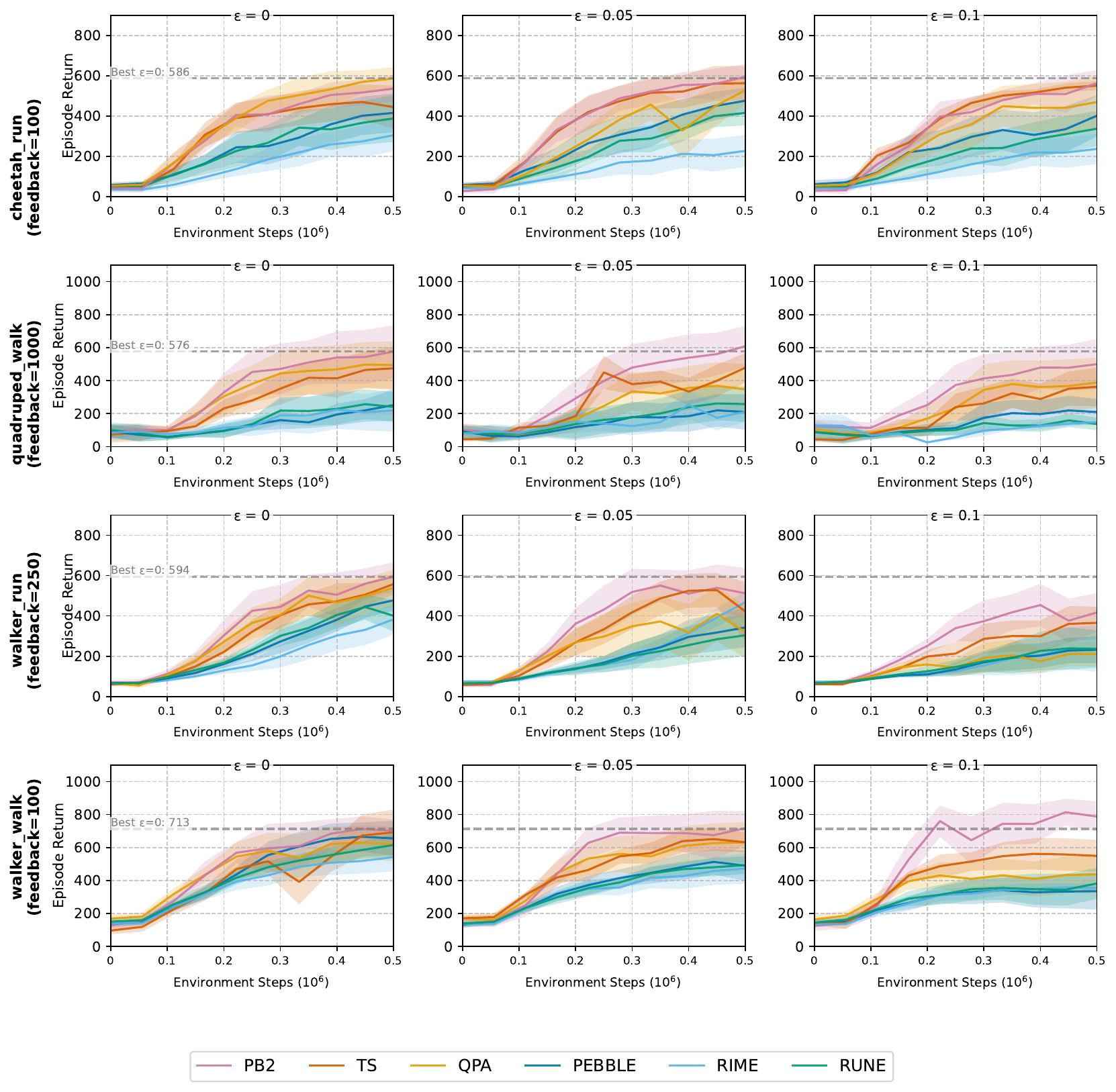}
    \caption{\textbf{Performance comparison on DMControl tasks with varying similarity thresholds.} Each column is a similarity threshold $\epsilon$ controlling when close trajectories yield inconsistent feedback. As $\epsilon$ increases from 0 (perfect oracle) to 0.1, PB² (pink) stays robust while the single-agent and naive posterior-sampling baselines degrade more, notably on quadruped\_walk and walker\_walk. Shaded regions denote one standard deviation over 10 seeds.}
    \label{fig:dmc_exp}
\end{figure}

\textbf{Escaping Local Optima in Preference Landscapes}
Fig.\ref{fig:qpa_failure} illustrates a failure case for single-agent methods. Starting from the same initial feedback (4 queries), after 20 feedback instances QPA remains trapped in a suboptimal region while PB² discovers a path to the high-reward region through its diverse population. This shows how population-based methods can escape local optima that trap single-agent approaches. The complete trajectory progression is in the appendix.

\textbf{Preference Exploration Improves Feedback Efficiency}
Table \ref{tab:nav2d_feedback} reports performance at varying feedback amounts. PB² outperforms QPA by up to 50\% in 2D Navigation with limited feedback (N=4 to N=8, narrowing at N=10), and by 20-30\% at most PointMaze budgets (N=12 to N=20). Population diversity thus enables more efficient exploration when feedback is scarce, which matters when minimizing human interaction.

\begin{table}
\centering
\caption{Performance comparison in 2D Navigation and PointMaze environments by feedback amount (N). Values are mean $\pm$ one standard deviation over 5 seeds.}
\scriptsize
\begin{tabular}{clcccc}
\toprule
& & \multicolumn{4}{c}{\textbf{Feedback Amount}}\\
\cline{3-6}\\
& \textbf{Algorithm} & \textbf{N=4} & \textbf{N=6} & \textbf{N=8} & \textbf{N=10} \\
\midrule
\multirow{3}{*}{\rotatebox[origin=c]{90}{\tiny{\textbf{2D Navig.}}}} & PB² & $\mathbf{-211.3 \pm 58.7}$ & $\mathbf{-141.0 \pm 48.5}$ & $\mathbf{-178.7 \pm 70.7}$ & $-126.5 \pm 18.9$ \\
&QPA & $-405.7 \pm 164.6$ & $-285.4 \pm 160.3$ & $-234.8 \pm 154.9$ & $\mathbf{-112.1 \pm 24.4}$\\
&PEBBLE & $-323.6 \pm 66.0$ & $-156.1 \pm 41.2$ & $-196.3\pm 95.9$ & $-164.5 \pm 71.1$ \\
&RUNE & $-359.8 \pm 78.5$ & $-204.6 \pm 109.6$ & $-251.45\pm 81.5$ & $-147.08 \pm 62.6$ \\
&TS & $-320.5 \pm 95.1$ & $-167.9 \pm 45.6$ & $-214.3 \pm 78.8$ & $-115.9 \pm 14.0$ \\
\midrule
& &  \textbf{N=8} & \textbf{N=12} & \textbf{N=16} & \textbf{N=20} \\
\midrule
\multirow{3}{*}{\rotatebox[origin=c]{90}{\tiny{\textbf{PointMaze}}}} & PB² & $18.9 \pm 3.63$                    & $\mathbf{85.0 \pm 45.3}$           & $\mathbf{132.2 \pm 28.7}$                    & $\mathbf{146.1 \pm 29.3}$ \\
&QPA & $30.7 \pm 37.0$          & $63.6 \pm 55.0$                    & $116.4 \pm 18.5$                             & $110.6 \pm 14.2$ \\
&PEBBLE & $\mathbf{33.2 \pm 30.5}$          & $80.6 \pm 63.1$                    & $91.4 \pm 60.4$                             & $98.1 \pm 52.5$ \\
&RUNE & $36.3 \pm 36.3$ & $57.9 \pm 38.3$ & $65.5\pm 38.6$ & $90.2 \pm 49.9$ \\
&TS & $30.3 \pm 57.3$ & $72.7\pm 26.3$ & $117.5\pm 70.6$ & $107.2 \pm 75.2$ \\
\bottomrule
\end{tabular}
\label{tab:nav2d_feedback}
\end{table}

\subsection{Evaluating distinguishability in high-dimensional continuous control}
\label{sec:distinguishability}

\textbf{Robustness to Trajectory Distinguishability Challenges.}
Fig.\ref{fig:dmc_exp} compares performance across DMControl tasks for $\epsilon \in \{0, 0.05, 0.1\}$. At $\epsilon=0$ (oracle teacher), PB² and QPA are comparable, but as $\epsilon$ increases PB² maintains stronger performance, particularly in quadruped\_walk and walker\_walk; on the latter at $\epsilon=0.1$ it reaches a return of about 750 versus around 400 for QPA and 350 for PEBBLE. By presenting more distinguishable segment pairs, PB² makes it easier for humans to give consistent feedback when trajectories are otherwise hard to differentiate.

Interestingly, the diversity bonus is central to this robustness. Despite using multiple reward functions and agents, the TS baseline (Sec.~\ref{sec:motiv}) performs very similarly to QPA, reinforcing our hypothesis that a vanilla neural ensemble does not represent the reward posterior any better than a single maximum-likelihood reward.

\section{Limitations and Future Work}

\textbf{Scalability with Population Size.}
Our implementation uses a small population (3 agents). With feedback budgets limited to approximately 10 evaluations per iteration, larger populations reduce learning iterations. Our analysis (Appendix Fig. \ref{fig:population}) shows 3-4 agents achieve optimal performance, with diminishing returns beyond that. Future work could explore adaptive population sizing that balances diversity against feedback constraints.

\textbf{Exploration-Exploitation Tradeoff hyperparameter.} PB² introduces $\lambda$ to control the balance between diversity and reward optimization. Our sensitivity analysis (Appendix Fig. \ref{fig:alpha}) shows moderate values perform well while extreme values converge to baseline performance. The performance-constrained mechanism (Equation 3) guarantees at least single-agent performance regardless of $\lambda$. We kept $\lambda$ consistent within each environment class rather than tuning per environment; individually tuning $\lambda$ could yield better performance. Future work could develop adaptive methods that automatically adjust this tradeoff, eliminating domain-specific tuning.

\section{Conclusion}

We presented PB², a population-based approach for preference-based reinforcement learning that addresses preference space exploration. By maintaining diverse agents, PB² generates more distinguishable behaviors for human evaluation. Our experiments confirm improved feedback efficiency, greater robustness to labeling inconsistency, and the ability to escape local optima, making PB² practical when human feedback is costly and potentially inconsistent.

\section*{Acknowledgments}
A. Davey and B. Driss were funded by the project ANR-23-CE23-0006. This work was granted access to the HPC resources of IDRIS under the allocation 2024-AD011015599 made by GENCI.

\clearpage

\bibliography{main}
\bibliographystyle{rlj}

\beginSupplementaryMaterials

\appendix

\section{Implementation Details}
\label{app:implementation}

\subsection{PB² Algorithm}
In this section, we provide the full procedure for PB², our population-based approach for preference-based reinforcement learning, in Algorithm~\ref{alg:pb2-full}.

\begin{algorithm}[h]
\caption{PB²: Population-Based Preference-Based Reinforcement Learning (Detailed)}
\label{alg:pb2-full}
\begin{algorithmic}[1]
\STATE {\bfseries Initialize:} Anchor policy $\pi_{\text{ref}}$, diverse policies $\{\pi_i\}_{i=2}^N$, discriminator $q_\psi$, reward model $r_\phi$
\STATE Initialize replay buffers $\{B_i\}_{i=1}^N \leftarrow \emptyset$ for each policy
\STATE Initialize preference dataset $D \leftarrow \emptyset$
\FOR{each unsupervised pre-training step $t$}
  \FOR{each policy $\pi_i$ in population}
      \STATE Collect $s_{t+1}^i$ by taking $a_t^i \sim \pi_i(a_t|s_t^i)$
      \STATE Compute state entropy reward $r_{\text{int}}^i \leftarrow -\log p(s_{t+1}^i)$
      \STATE Store transitions $B_i \leftarrow B_i \cup (s_t^i, a_t^i, s_{t+1}^i, r_{\text{int}}^i)$
  \ENDFOR
  \FOR{each gradient step}
      \FOR{each policy $\pi_i$ in population}
          \STATE Sample minibatch $(s_j^i, a_j^i, s_{j+1}^i, r_{\text{int},j}^i)_{j=1}^B \sim B_i$
          \STATE Optimize policy $\pi_i$ using SAC with intrinsic reward
      \ENDFOR
  \ENDFOR
\ENDFOR
\WHILE{feedback budget not exhausted}
  \STATE {\bfseries // Experience Collection Phase}
  \FOR{each environment step}
      \FOR{each policy $\pi_i$ in population}
          \STATE Collect $s_{t+1}^i$ by taking $a_t^i \sim \pi_i(a_t|s_t^i)$
          \STATE Compute reward $\hat{r}_t^i = r_\phi(s_t^i, a_t^i)$
          \STATE Compute discriminator reward $r_{\text{disc}}^i = \log q_\psi(i|s_t^i) - \log p(i)$
          \STATE Store transitions $B_i \leftarrow B_i \cup (s_t^i, a_t^i, s_{t+1}^i, \hat{r}_t^i, r_{\text{disc}}^i)$
      \ENDFOR
  \ENDFOR

  \STATE {\bfseries // Feedback Collection Phase}
  \IF{step to query preferences}
      \STATE Sample $K$ recent trajectories from each policy's replay buffer $\{B_i\}_{i=1}^N$
      \STATE Randomly select trajectory segments to form candidate query set $\mathcal{Q} = \{(\sigma_0, \sigma_1)\}$
      \STATE Collect human feedback $\{y_i\}$ for queries in $\mathcal{Q}$
      \STATE Store preferences $D \leftarrow D \cup \{(\sigma_0^i, \sigma_1^i, y_i)\}$

      \STATE {\bfseries // Reward Model Update}
      \STATE Update reward model $r_\phi$ using dataset $D$ by minimizing loss in Equation (2)
      \STATE Relabel replay buffers $\{B_i\}_{i=1}^N$ using updated $r_\phi$
  \ENDIF

  \STATE {\bfseries // Policy Optimization Phase}
  \STATE {\bfseries // Anchor Agent Update}
  \STATE Calculate anchor performance $R_{\text{ref}} = \mathbb{E}_{\tau \sim \pi_{\text{ref}}}[R_\phi(\tau)]$
  \STATE Update anchor policy $\pi_{\text{ref}}$ to maximize $\mathbb{E}[r_\phi(s, a)]$ using SAC

  \STATE {\bfseries // Diverse Agents Update}
  \FOR{each diverse policy $\pi_i, i \geq 2$}
      \STATE Calculate agent performance $R_i = \mathbb{E}_{\tau \sim \pi_i}[R_\phi(\tau)]$
      \IF{$R_i \geq \alpha \cdot R_{\text{ref}}$}
          \STATE Update $\pi_i$ to maximize $\mathbb{E}[r_\phi(s, a) + \lambda \cdot r_{\text{disc}}^i(s)]$ using SAC
      \ELSE
          \STATE Update $\pi_i$ to maximize $\mathbb{E}[r_\phi(s, a)]$ using SAC (no diversity bonus)
      \ENDIF
  \ENDFOR

  \STATE {\bfseries // Discriminator Update}
  \STATE Collect state samples $\{s_j^i\}$ from each policy $\pi_i$
  \STATE Update discriminator $q_\psi$ by maximizing $\mathbb{E}[\log q_\psi(i|s_j^i)]$
\ENDWHILE
\end{algorithmic}
\end{algorithm}

\subsection{Population Management}

\subsubsection{Population Design choice}

Unlike methods such as DIAYN and SMERL that use a single policy network conditioned on a latent variable to learn multiple behaviors, PB² maintains separate policy networks for each agent in the population. This design choice offers key advantages: it allows for independent update of population members, enables different agents to fulfill distinct roles (exploration vs. exploitation) and provides a robust mechanism for recovery when individual agents fail. Having a dedicated anchor that focuses solely on maximizing the learned reward function creates a stable anchor for the population, serving as a reliable fallback that can replace underperforming explorer agents when necessary.

\subsubsection{Reference Agent Approach}

One of the key innovations in PB² is our anchor agent mechanism, which provides a stable performance benchmark without requiring access to an expert policy or ground-truth reward function. This section details the complete implementation of this approach.

The anchor (indexed as agent 0 in our implementation) is trained to maximize only the current reward model without any diversity bonus:

\begin{equation}
\pi_0 \in \arg\max_{\pi} \mathbb{E}_{\tau \sim \pi} \left[ \sum_{t=0}^{T} \gamma^t r_\phi(s_t, a_t) \right]
\end{equation}

The reward model $r_\phi$ is continuously updated based on human preference feedback. After each reward model update, we track the anchor's expected return under the new reward function:

\begin{equation}
R_{\text{ref}} = \mathbb{E}_{\tau \sim \pi_0} \left[ \sum_{t=0}^{T} \gamma^t r_\phi(s_t, a_t) \right]
\end{equation}

In practice, this expectation is approximated by maintaining a running average of episode returns for the anchor over a window of recent episodes.

The other agents in the population (indexed 1 through $N - 1$) are trained with the following objective:

\begin{equation}
\pi_i \in \arg\max_{\pi} \mathbb{E}_{\tau \sim \pi} \left[ \sum_{t=0}^{T} \gamma^t \left( r_\phi(s_t, a_t) + \lambda \cdot \mathds{1}_{[R_\phi(\tau) \geq \alpha \cdot R_{\text{ref}}]} \cdot \log q_\psi(i|s_t) \right) \right]
\end{equation}

Where:
\begin{itemize}
    \item $\lambda$ is the diversity coefficient (typically 0.25 for locomotion tasks, 0.5 for navigation)
    \item $\alpha$ is the performance threshold (0.9 in our implementation)
    \item $\mathds{1}_{[R_\phi(\tau) \geq \alpha \cdot R_{\text{ref}}]}$ is an indicator function that equals 1 when the agent's expected return is at least $\alpha$ times the anchor return
    \item $q_\psi(i|s)$ is the discriminator that predicts which agent generated a given state
\end{itemize}

In practice, the discriminator $q_\psi(i|s)$ is implemented as a neural network that takes states as input and outputs a probability distribution over the $N$ agents. The diversity reward $\log q_\psi(i|s)$ is computed for each state encountered during training and added to the extrinsic reward $r_\phi(s,a)$ when the performance constraint is satisfied. In our implementation, we track the recent performance of each agent and apply the diversity bonus only when an agent's performance exceeds the threshold relative to the anchor. This performance-constrained diversity mechanism ensures that exploration through diversity is only encouraged when it doesn't significantly compromise task performance.

\subsubsection{Policy Inheritance in the Population}

After each feedback session in PB², we transfer the critic and actor networks from the anchor to all explorer agents in the population. This inheritance mechanism preserves the core task knowledge while still enabling agent specialization through the diversity bonuses. It serves multiple purposes:

It allows all agents to quickly benefit from updated reward models based on human feedback, ensuring the entire population adapts to new preference information simultaneously. It maintains a healthy balance between exploration and exploitation across the population, with the anchor agent focusing on exploitation while explorer agents develop diverse behaviors from a common foundation. Most importantly, it prevents explorer agents from drifting too far from the task objective due to diversity pressures, which could otherwise lead to behaviors that are novel but ineffective.

This periodic knowledge sharing proved particularly beneficial in complex environments where unguided exploration can be inefficient, allowing the population to maintain both performance and diversity throughout training.

\subsection{Stability Mechanisms}

\subsubsection{Transition Handling Between Preference Updates}

A critical challenge in our population-based approach is managing the discriminator's behavior during preference transitions. When the reward model updates, we implement several specific techniques to ensure stable and effective diversity guidance:

\begin{enumerate}
    \item \textbf{Temporary Diversity Suspension:} Immediately after a reward model update, we temporarily disable the diversity bonus by setting the discriminator coefficient to zero for a fixed period (typically 5000 environment steps). This suspension period allows all agents to first adapt to the new reward landscape before reintroducing diversity incentives, preventing potentially counterproductive exploration based on outdated preferences.

    \item \textbf{On-Policy Discriminator Training:} During transition periods, the discriminator is trained exclusively on recent experiences rather than the full replay buffer. This is implemented through our on-policy sampling mechanism that selects state samples from only the most recent trajectories for each agent.

    \item \textbf{Reward Normalization and Clipping:} To prevent extreme diversity bonuses during transition periods, we implement Exponential Moving Average (EMA) normalization with a decay rate of 0.99 for the discriminator's outputs. Additionally, we clip the normalized intrinsic rewards to the range $[-2, 2]$, which prevents destabilizing spikes in the diversity bonus that could otherwise derail learning during preference transitions.
\end{enumerate}

These implementation details are crucial for maintaining stable diversity during preference transitions, as they address the practical challenges of keeping a population-based system aligned with changing preference signals. By temporarily reducing diversity pressure, carefully normalizing rewards, and ensuring balanced training, we allow the discriminator to smoothly adapt to new preference landscapes without destabilizing the learning process.

\section{Technical Details}

\subsection{Implementation Framework}
\label{app:implementation_framework}

Our implementation builds upon established preference-based reinforcement learning frameworks. We extend the QPA codebase \citep{Nakamoto2023QPA}, which itself extends the B-Pref framework \citep{Lee2021a} that provides core functionality for preference-based learning through the PEBBLE \citep{Lee2021b} algorithm architecture.

To accommodate our population-based approach, we expanded this framework to manage multiple policy networks simultaneously, while incorporating the on-policy query selection mechanism introduced in QPA \citep{Nakamoto2023QPA}. This combination allows us to leverage on-policy query selection across a diverse population of agents, rather than limiting it to a single policy.

Each agent in our population maintains its own replay buffer for experience collection and policy optimization. We introduced additional components for our discrimination-based diversity mechanism, and policy inheritance procedures.

For comparisons against baselines, we integrated implementations of competing methods including PEBBLE \citep{Lee2021b}, RUNE \citep{Wang2023RUNE} and QPA \citep{Nakamoto2023QPA} using their official codebases, ensuring fair evaluation across all approaches. Our implementation maintains compatibility with the underlying frameworks while introducing the population management and diversity mechanisms that define PB².

\subsection{Network Architectures}

For fair comparison, PB² uses the same network architectures as QPA for both the SAC algorithm and reward model, with our modifications focused exclusively on the population-based exploration mechanism.

The reward model consists of 3 hidden layers with 256 units each and LeakyReLU activations, taking state-action pairs as input and producing scalar reward predictions with tanh activation.

The policy networks follow the standard SAC architecture with 2 hidden layers of 1024 units and ReLU activations. The actor outputs action means and log standard deviations (bounded between -5 and 2), while the critic takes state-action pairs as input and outputs Q-value predictions.

Our discriminator network uses 2 hidden layers of 256 units with ReLU activations and layer normalization. The network takes states $s \in \mathbb{R}^{d_s}$ as input and outputs a probability distribution over the $N$ agents via softmax activation. Formally:

\begin{equation}
q_\psi(i|s) = \frac{\exp(f_\psi(s)_i)}{\sum_{j=1}^{N} \exp(f_\psi(s)_j)}
\end{equation}

where $f_\psi: \mathbb{R}^{d_s} \to \mathbb{R}^N$ is the discriminator network. This discriminator is trained using cross-entropy loss on balanced batches of states from each policy in the population.

\subsection{Hyperparameters}

We maintain consistent hyperparameters across all comparison methods for fair evaluation, with the only differences being in the population-specific parameters introduced by PB². Tables \ref{tab:general_hyperparameters}-\ref{tab:environment_hyperparameters} provide a comprehensive overview of the hyperparameters used in our experiments.

The general hyperparameters (Table \ref{tab:general_hyperparameters}) are common across all environments, while environment-specific parameters (Table \ref{tab:environment_hyperparameters}) highlight the varying complexity and requirements of different tasks. DMControl tasks generally required more feedback than navigation tasks due to their higher-dimensional state and action spaces, while maintaining a consistent diversity coefficient of 0.25. Navigation tasks benefited from a higher diversity coefficient (0.5) to encourage more extensive exploration of the state space.

For the discriminator-related parameters (Table \ref{tab:discriminator_hyperparameters}), we performed a limited hyperparameter search and found that a learning rate of 1e-5 and hidden size of 256 worked well across environments. The reward model (Table \ref{tab:reward_hyperparameters}) and SAC parameters (Table \ref{tab:sac_hyperparameters}) were selected to match those used in prior work for direct comparison.

\begin{table}[h]
\centering
\begin{tabular}{lll}
\hline
\textbf{Parameter} & \textbf{Description} & \textbf{Value} \\
\hline
Discount factor ($\gamma$) & Reward discount factor & 0.99 \\
Replay buffer capacity & Maximum transitions stored & Training steps \\
Population size & Number of agents (including anchor) & 3 \\
Activation function & For all networks & tanh \\
Gradient update frequency & Updates per environment step & 1 \\
\hline
\end{tabular}
\caption{General hyperparameters used in the PB² algorithm}
\label{tab:general_hyperparameters}
\end{table}

\begin{table}[h]
\centering
\begin{tabular}{lll}
\hline
\textbf{Parameter} & \textbf{Description} & \textbf{Value} \\
\hline
Actor learning rate & Learning rate for policy network & 5e-4 \\
Critic learning rate & Learning rate for Q-networks & 5e-4 \\
Alpha learning rate & Learning rate for temperature parameter & 1e-4 \\
Initial temperature & Initial value of entropy coefficient & 0.1 \\
Target update rate ($\tau$) & Polyak averaging coefficient & 0.005 \\
Target update frequency & Steps between target network updates & 2 \\
Actor update frequency & Steps between policy updates & 1 \\
\hline
\end{tabular}
\caption{SAC hyperparameters used in the PB² algorithm}
\label{tab:sac_hyperparameters}
\end{table}

\begin{table}[H]
\centering
\begin{tabular}{lll}
\hline
\textbf{Parameter} & \textbf{Description} & \textbf{Value} \\
\hline
Ensemble size & Number of networks in reward ensemble & 1 \\
Learning rate & Learning rate for reward model & 3e-4 \\
Number of hidden layers & Hidden layers in reward network & 3 \\
Hidden size & Units per hidden layer & 256 \\
\hline
\end{tabular}
\caption{Reward model hyperparameters}
\label{tab:reward_hyperparameters}
\end{table}

\begin{table}[h]
\centering
\begin{tabular}{lll}
\hline
\textbf{Parameter} & \textbf{Description} & \textbf{Value} \\
\hline
Batch size & States per gradient update & 256 \\
Learning rate & Learning rate for discriminator & $1 \times 10^{-5}$ \\
Hidden size & Units per hidden layer & 256 \\
Activation & Hidden layer activation & ReLU \\
Normalization & Layer normalization & Yes \\
On-policy ratio & Ratio of on-policy samples for training & 0.5 \\
Reward clipping & Range for normalized rewards & $[-2, 2]$ \\
EMA decay & Decay rate for running statistics & 0.99 \\
\hline
\end{tabular}
\caption{Discriminator hyperparameters}
\label{tab:discriminator_hyperparameters}
\end{table}

\begin{table}[h]
\centering
\begin{tabular}{lllll}
\hline
\textbf{Environment} & \textbf{Feedback Budget} & \textbf{Total Steps} & \textbf{Diversity $\lambda$ } & \textbf{Other Parameters} \\
\hline
Cheetah\_run & 100 & 500,000 & 0.25 & Segment length: 50 \\
Walker\_walk & 100 & 500,000 & 0.25 & Unsupervised steps: 9,000 \\
Walker\_run & 250 & 500,000 & 0.25 & Interact steps: 20,000 \\
Quadruped\_walk & 1000 & 500,000 & 0.25 & Queries per iteration: 10 \\
\hline
2D Navigation & 10 & 20,000 & 0.5 & Segment length: 20-50 \\
PointMaze & 20 & 80,000 & 0.5 & Unsupervised steps: 900-400 \\
&  &  &  & Interact steps: 2,000-10,000 \\
&  &  &  & Queries per iteration: 2-4 \\
\hline
\end{tabular}
\caption{Environment-specific hyperparameters}
\label{tab:environment_hyperparameters}
\end{table}

\section{Evaluation Methodology Details}
\label{app:eval_methodology}
For simulating human evaluation inconsistency, we implement the Equal teacher from B-Pref \citep{Lee2021a} with a similarity threshold mechanism. Specifically, for a query comparing trajectories with ground truth returns $R_1$ and $R_2$, we assign random preference labels when $|R_1 - R_2| < \epsilon \cdot \max(R_1, R_2)$, where $\epsilon$ is the similarity threshold parameter. This approach only introduces inconsistency when trajectories are genuinely similar and difficult to distinguish, better simulating human evaluation challenges. While $\epsilon=0$ (no similarity threshold) provides a theoretical baseline, we emphasize results with $\epsilon>0$ as more realistic, since human evaluators inevitably provide less consistent feedback when comparing similar behaviors.

\section{Environment Details}
\label{app:environments}
To thoroughly evaluate our population-based approach for preference-based RL, we selected environments that specifically challenge the two key aspects of our method: efficient preference space exploration and robustness to human feedback inconsistency. Our environment selection targets two critical scenarios: (1) low-feedback settings where efficient exploration is crucial, and (2) complex environments where trajectory similarity makes human evaluation difficult and error-prone.

Navigation tasks provide intuitive visualization of exploration patterns and allow us to demonstrate how our method escapes local optima in preference landscapes with limited feedback. DMControl tasks, with their high-dimensional state-action spaces, create scenarios where trajectories can appear similar to human evaluators despite having different underlying rewards, testing our method's ability to generate distinguishable queries and handle inconsistent feedback.

While previous methods often include robotic manipulation tasks from Meta-World, these environments typically require substantially more feedback (often 2,000-3,000 queries) to achieve meaningful performance. Such high feedback requirements are unrealistic in practical human-in-the-loop scenarios. Nevertheless, we include one high-feedback experiment (1,000 queries) on the Quadruped\_walk task to demonstrate our method's performance across the feedback spectrum.

\subsection{Navigation Tasks}

\subsubsection{2D Navigation}
The 2D navigation environment consists of a $10 \times 10$ continuous arena where the agent starts at the bottom left corner $(0,0)$ and must navigate to a goal position at the top right corner $(10,10)$. The state space is 2-dimensional, corresponding to the agent's $(x,y)$ position. The action space is also 2-dimensional, where actions directly change the agent's position with values in the range $[-1, 1]$. If the agent attempts to move outside the boundaries of the arena, it is projected to the closest point inside. The reward function used for ground truth evaluation (not accessible to the agent) is the negative Euclidean distance to the goal position.

For human feedback simulation, we compare trajectory segments of length 50 timesteps. The oracle provides preferences based on the total progress made toward the goal during each segment. When the similarity threshold $\epsilon$ is applied, random labels are provided when the difference in progress between segments falls below the threshold.

\subsubsection{PointMaze}
In the PointMaze environment, a point mass agent navigates through a maze with walls to reach a designated goal location. The state space consists of the agent's position and velocity (4D). The action space is 2-dimensional, controlling the force applied in the $x$ and $y$ directions.

Instead of using the original reward function based on Euclidean distance to the goal, we replace it with a handcrafted reward function that better aligns with human preferences by guiding the agent through the maze as shown in Fig.\ref{fig:qpa_failure}. This reward function provides higher values along the correct path through the maze corridors, creating a more structured reward landscape that captures the preference for following the intended route rather than attempting to move directly toward the goal (which would cause collisions with walls). This modification helps simulate realistic human preferences that incorporate domain knowledge about the maze's structure rather than simple distance metrics.

\subsection{DMControl Tasks}

We evaluate our method on four continuous control tasks from the DeepMind Control Suite (DMControl) \citep{tassa2018deepmind}: Cheetah\_run, Walker\_run, Walker\_walk, and Quadruped\_walk. These environments feature continuous state and action spaces with increasing complexity, from the 17-dimensional state space of Cheetah\_run to the 78-dimensional state space of Quadruped\_walk. All DMControl tasks have episode lengths of 1000 timesteps.

The ground truth rewards in these environments, which are used only for evaluation and not accessible to the learning algorithms, combine task-specific objectives (such as forward velocity above environment-specific thresholds) with control penalties. The Walker and Quadruped environments additionally reward upright posture maintenance.

\section{Additional Experimental Results}

\subsection{Component Ablation Study}

\begin{figure}[t]
    \centering
    \includegraphics[width=0.8\linewidth]{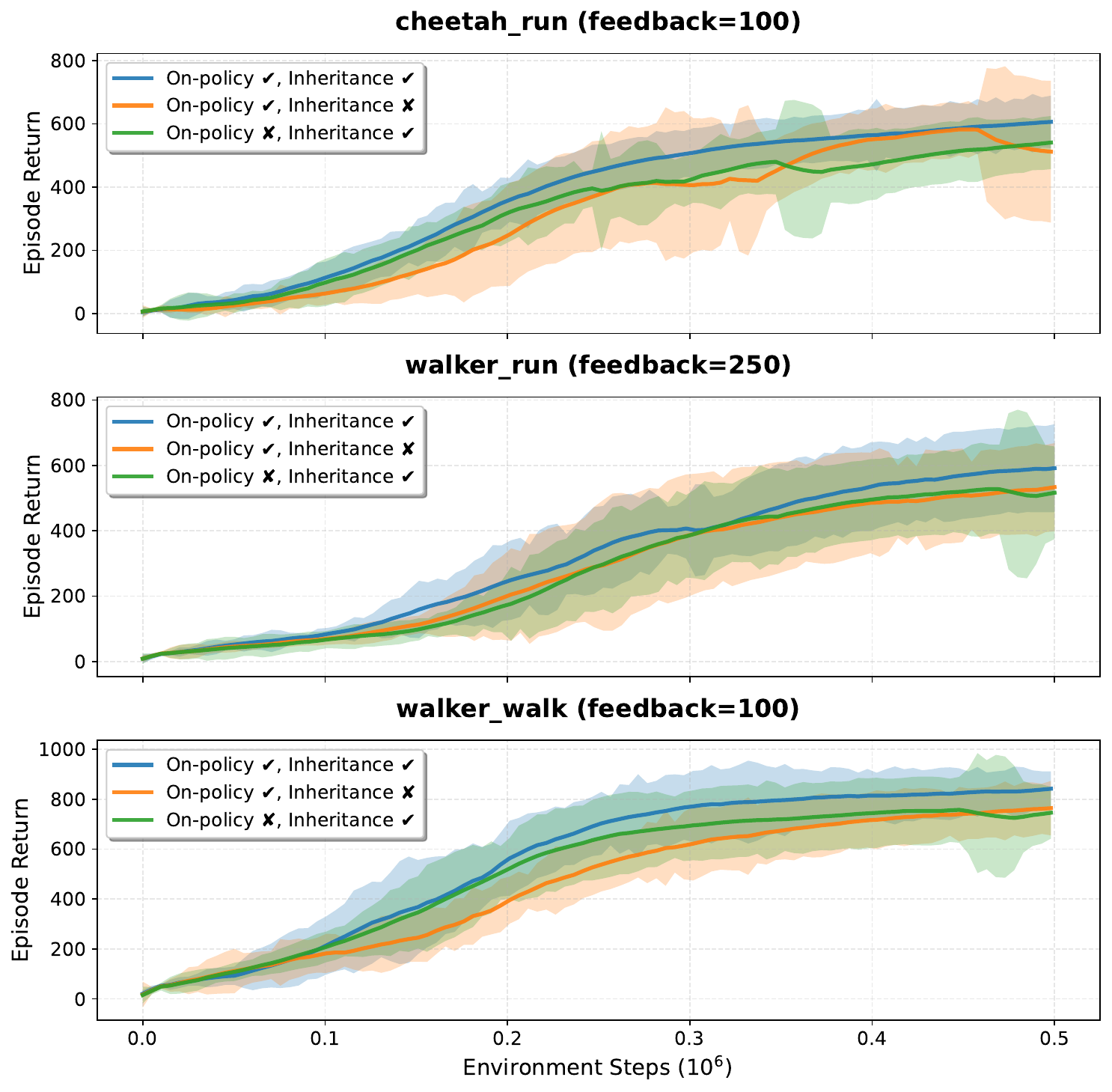}
    \caption{Ablation study on DMControl locomotion tasks showing the contribution of key components in PB². Results demonstrate that both on-policy query generation and agent inheritance mechanisms are essential for achieving optimal performance across different environments and feedback budgets.}
    \label{fig:ablation}
\end{figure}

To evaluate the contribution of individual components in PB², we conducted ablation studies examining the impact of on-policy sampling and policy inheritance mechanisms. Figure~\ref{fig:ablation} shows results across three DMControl environments.

The results demonstrate that both on-policy sampling and inheritance contribute positively to performance. The full PB² method (On-policy \checkmark, Inheritance \checkmark) consistently achieves the highest performance across all environments. Removing policy inheritance (On-policy \checkmark, Inheritance $\times$) leads to noticeable performance degradation, particularly in the later stages of training. This confirms the importance of knowledge sharing between agents in the population. Interestingly, removing on-policy sampling while keeping inheritance (On-policy $\times$, Inheritance \checkmark) shows competitive performance in some environments, suggesting that the benefits of these components can be complementary depending on the task complexity.

\subsection{Diversity Parameter Analysis}
\begin{figure}[t]
    \centering
    \includegraphics[width=0.8\linewidth]{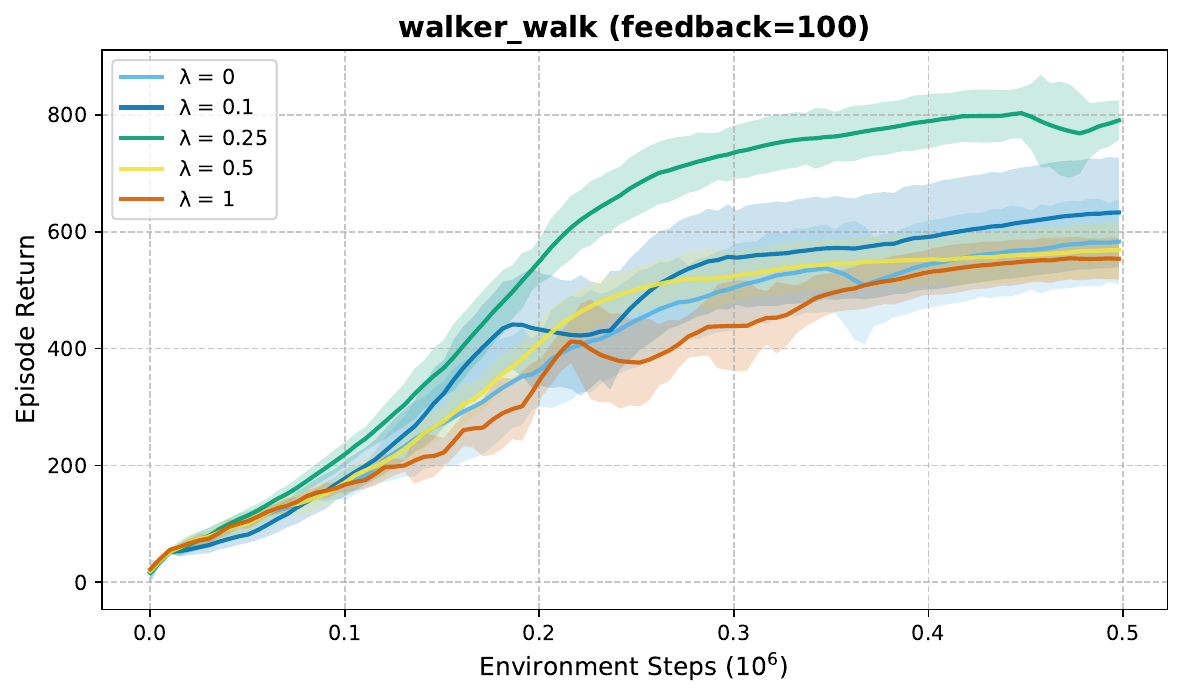}
    \caption{Sensitivity analysis of the diversity parameter $\lambda$ in the walker\_walk environment with 100 feedback queries. The results show that small values of $\lambda$ (0.1-0.25) achieve the best balance between exploration and exploitation in this setup, while extreme values ($\lambda = 0$ or $\lambda = 1$) lead to suboptimal performance.}
    \label{fig:alpha}
\end{figure}

We investigated the effect of the diversity parameter $\lambda$ on learning performance. Figure~\ref{fig:alpha} shows results for the Walker\_walk environment with different values of $\lambda$ ranging from 0 to 1.

The results reveal that moderate values of $\lambda$ (0.1-0.25) achieve the best performance, with $\lambda$ = 0.25 showing the strongest results. Setting $\lambda$ = 0 (no diversity bonus) leads to reduced performance due to insufficient exploration and limited behavioral diversity across the population. Conversely, very high values ($\lambda$ = 1) also underperform, suggesting that excessive emphasis on diversity can distract agents from optimizing the primary reward signal. The optimal range around $\lambda$ = 0.25 (for the current environment) provides an effective balance between reward maximization and exploration diversity, enabling agents to discover distinct yet effective behaviors.

\subsection{Population Size Sensitivity}
\begin{figure}[t]
    \centering
    \includegraphics[width=0.8\linewidth]{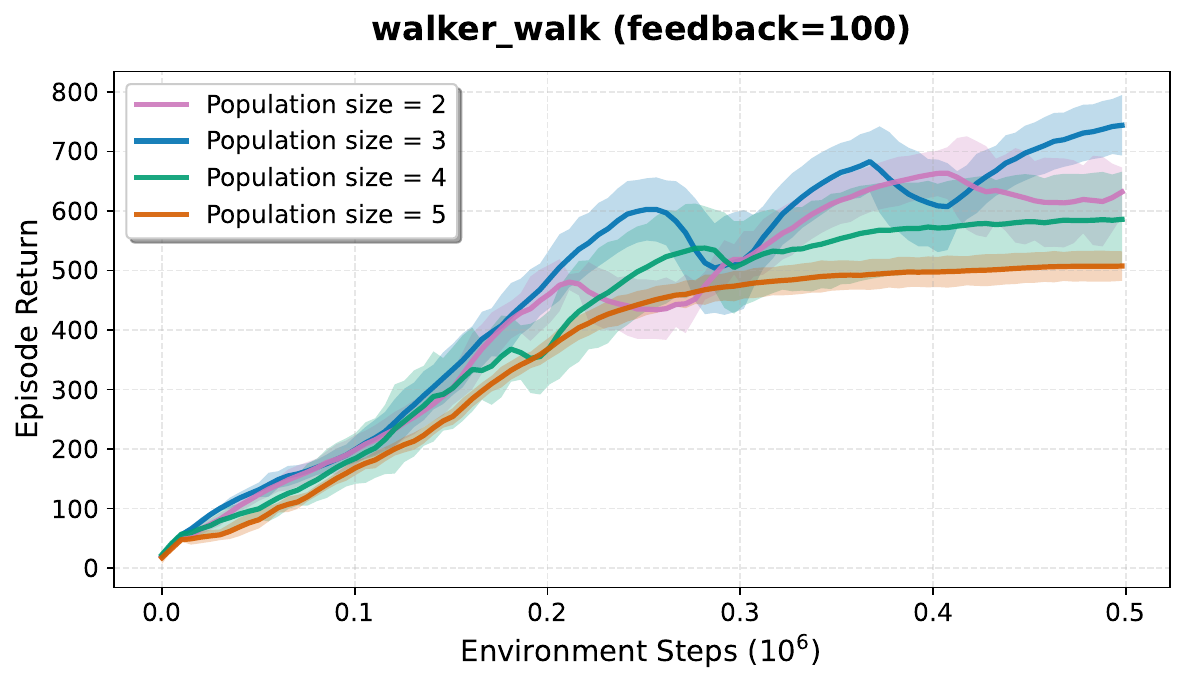}
    \caption{Impact of population size on learning performance in the walker\_walk environment with 100 feedback queries. Results indicate that a population size of 3-4 agents provides the optimal trade-off between diversity benefits and computational efficiency, with diminishing returns for larger populations.}
    \label{fig:population}
\end{figure}

Figure~\ref{fig:population} examines the impact of different population sizes (2, 3, 4, 5 agents) on learning performance in the Walker\_walk environment.

The results show that population sizes of 3-4 agents achieve the best performance, with diminishing returns as population size increases further. A population of size 2 shows competitive early performance but fails to maintain the same final performance level as larger populations. This is likely because with a fixed query budget of 10 per iteration, having only 2 agents results in trajectories that are too similar to each other, limiting the diversity of preference queries. Conversely, populations of size 5 do not provide significant benefits over size 4, potentially because the available queries become too sparse across agents, reducing the learning signal for each individual policy. The population size of 3, which we use in our main experiments, appears to be well-chosen based on this analysis, providing sufficient diversity while maintaining concentrated learning signals.

\subsection{Anchor Threshold Sensitivity}

\begin{figure}[t]
    \centering
    \includegraphics[width=0.8\linewidth]{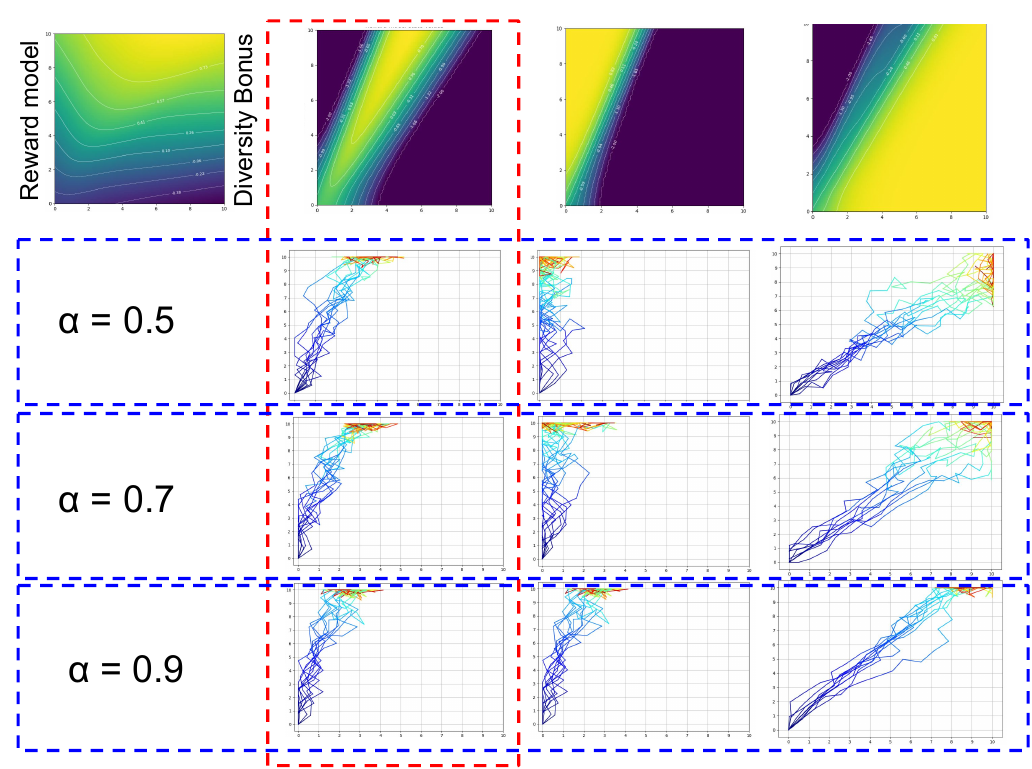}
    \caption{Anchor threshold sensitivity analysis showing the effect of $\alpha$ on multi-objective optimization in a 2D navigation task. The parameter $\alpha$ sets the performance threshold relative to the anchor agent that diverse agents must meet to receive diversity bonuses. Higher $\alpha$ values require stricter reward model alignment before agents can optimize for both reward model predictions and diversity bonuses simultaneously, leading to trajectories that target the intersection of high reward and high diversity regions. The top row shows reward model predictions and individual diversity bonuses for three agents, while trajectories demonstrate how different $\alpha$ values affect agents' ability to balance both objectives.}
    \label{fig:alpha_ablation}
\end{figure}

The anchor threshold parameter $\alpha$ controls when diverse agents receive the diversity bonus by setting a performance threshold relative to the anchor agent. Figure~\ref{fig:alpha_ablation} demonstrates how different values of $\alpha$ (0.5, 0.7, 0.9) affect agent behavior in a 2D navigation environment. Higher $\alpha$ values create stricter conditions for agents to receive the diversity bonus, requiring them to achieve higher performance relative to the anchor before being allowed to optimize for both the reward model and diversity bonus simultaneously. When agents meet this threshold, their trajectories aim for regions that are high in both the reward model predictions (shown in yellow in the leftmost plot) and their individual diversity bonuses. With $\alpha$ = 0.5, the lenient threshold allows agents to quickly access diversity bonuses, leading to trajectories that target the intersection of high reward and high diversity regions. At $\alpha$ = 0.7, agents show more selective alignment to these intersection regions. At $\alpha$ = 0.9, the strict performance requirement means agents must demonstrate strong reward model alignment before they can optimize for the combination of both rewards, resulting in trajectories that more carefully target regions where both the reward model and diversity bonus are simultaneously high. The diversity bonus visualizations (top row) show distinct high-value regions for each agent, and the trajectories demonstrate how agents navigate toward areas where these bonuses overlap with reward model predictions.

\subsection{Detailed Trajectory Evolution in Point Maze Environment}

This section provides the complete trajectory progression referenced in Fig.4 of the main text, showing the step-by-step evolution of exploration patterns as feedback increases from N=4 to N=16 queries in the Point Maze environment.

\begin{figure}[t]
\centering
\includegraphics[width=\textwidth]{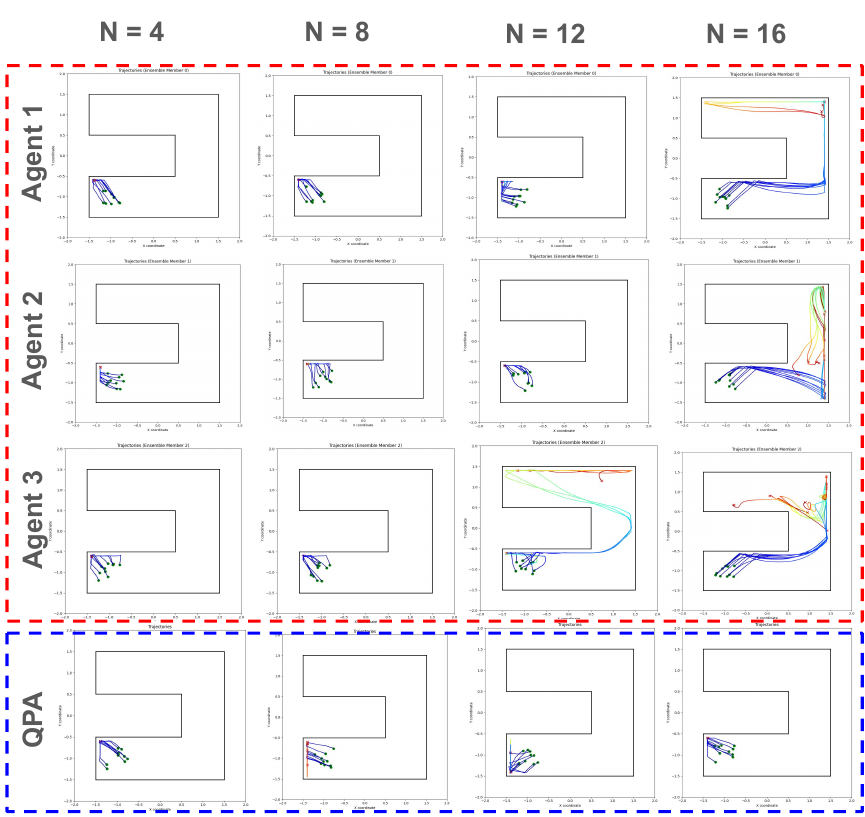}
\caption{Complete trajectory evolution in Point Maze environment showing exploration patterns of PB² (red box, with Agent 1, 2, 3) versus QPA (blue box) as feedback increases from N=4 to N=16 queries. The progression demonstrates how PB²'s population-based approach maintains diverse exploration strategies across multiple agents, while QPA's single-agent method becomes trapped in suboptimal regions due to its exploitative nature.}
\label{fig:maze_progression}
\end{figure}

Figure~\ref{fig:maze_progression} illustrates how PB² (red box) and QPA (blue box) evolve their exploration strategies with increasing feedback. At N=4, both methods show similar initial exploration patterns after receiving identical feedback. However, as feedback increases, the population-based approach in PB² enables the three agents to explore distinct regions of the maze, attempting to find high reward regions. In contrast, QPA's single-agent approach becomes increasingly concentrated in the initially promising but suboptimal upper-left region, demonstrating the local optima problem discussed in the main text.

Crucially, while not all agents in PB² necessarily discover the optimal path simultaneously, once one agent finds a better trajectory (Agent 3, N=12), the reward model update incorporates this improved knowledge by comparing it against the previous suboptimal behaviors from other agents. This allows the shared reward model to capture the superior strategy, subsequently guiding all agents toward the newly discovered high reward region (N=16, Agents 1,2 and 3). This progression clearly shows how PB²'s diverse population prevents premature convergence and enables discovery of multiple pathways that eventually lead to finding the optimal solution, while QPA remains trapped in its initial exploration pattern.

\section{Reproducibility}
\label{app:compute}


\paragraph{Code Repository}
Our implementation is publicly available at \url{https://github.com/brahimdriss/PB2_RLC}. The repository includes the complete codebase with all experimental configurations, training scripts, and evaluation utilities necessary to reproduce our results.

\end{document}